\documentclass{article} % For LaTeX2e
\usepackage{iclr2026_conference,times}

\usepackage{amsmath,amsfonts,bm}

\def\eqref#1{equation~\ref{#1}}
\def\1{\bm{1}}

\DeclareMathAlphabet{\mathsfit}{\encodingdefault}{\sfdefault}{m}{sl}
\SetMathAlphabet{\mathsfit}{bold}{\encodingdefault}{\sfdefault}{bx}{n}

\usepackage{hyperref}
\usepackage{url}

\usepackage{algorithm}
\usepackage{algpseudocode}
\usepackage{graphicx}

\title{Chebyshev-Augmented One-Shot Transfer Learning for PINNs on Nonlinear Differential Equations}

\author{Yiqi Rao \\
Harvard University \\
\texttt{herryrao@g.harvard.edu} \\
\And
Pavlos Protopapas \\
Harvard University \\
\texttt{pprotopapas@g.harvard.edu}
}

\iclrfinalcopy % Uncomment for camera-ready version, but NOT for submission.
\begin{document}

\maketitle

\begin{abstract}
Physics-Informed Neural Networks (PINNs) offer a flexible paradigm for solving differential equations by embedding governing laws into the training objective. A persistent limitation is instance specificity: standard PINNs typically require retraining for each new forcing term, boundary/initial condition, or parameter setting. One-shot transfer learning (OTL) addresses this bottleneck for linear operators by freezing a pretrained latent representation and computing optimal output weights in closed form, but for nonlinear problems closed-form adaptation is generally unavailable because the loss is nonconvex in the output layer.

In this paper we substantially broaden the class of nonlinearities amenable to one-shot PINN transfer by combining OTL with Chebyshev polynomial surrogates. We approximate general smooth weakly nonlinear terms by truncated Chebyshev expansions over a prescribed solution range, yielding a polynomial nonlinearity that can be handled by a perturbative decomposition into linear subproblems. A multi-head PINN learns a reusable latent space associated with the dominant linear operator; at test time, solutions to new instances are obtained via a sequence of closed-form linear solves in the output layer, without retraining the network body.

We provide a unified derivation of the framework for ODEs and PDEs and demonstrate accuracy and fast online adaptation on nonlinear benchmarks, including non-polynomial and singular ODE nonlinearities as well as a reaction–diffusion PDE with saturating kinetics, demonstrating the method’s utility in many-query regimes.
\end{abstract}

\section{Introduction}\label{sec:intro}

PINNs have become a widely used tool for solving forward and inverse differential-equation problems by minimizing residuals of the governing equations alongside boundary/initial conditions \citep{lagaris1998anns,raissi2019pinns,karniadakis2021piml,desai2022oneshot}. Despite their flexibility, a core drawback remains: they are typically trained separately for each problem instance. Even when two problems share the same operator but differ only in forcing or boundary data, standard PINN approaches generally require a new optimization run, limiting their utility in many-query, real-time, or interactive settings where rapid adaptation is required.

Transfer learning for PINNs aims to amortize training across a family of related problems. A particularly strong form is one-shot transfer learning (OTL): for linear ODE/PDE families, a shared network body can be frozen and the output-layer weights computed in closed form, yielding fast adaptation with a single matrix inversion \citep{desai2022oneshot,lei2023nonlinearodes}. However, the presence of nonlinear terms renders the optimization problem nonconvex in the output weights, precluding closed-form adaptation in general.

A promising recent direction is to restore linear solvability via perturbative decomposition. For nonlinearities that appear as a small polynomial perturbation, the nonlinear problem can be expressed as a sequence of linear subproblems whose right-hand sides depend on lower-order solutions \citep{lei2023nonlinearodes,auroy2025nonlinearpdes,alexandrino2026ptl}. Each linear subproblem can then be solved by one-shot weight updates. While effective, this framework is restricted to weak polynomial nonlinearities and to regimes where the perturbative expansion remains valid.

In this work, we propose to relax the restrictive polynomial assumption by approximating weak nonlinear terms with truncated Chebyshev polynomial expansions \citep{trefethen2019atapia,boyd2001cheb,mason2002chebyshev}. This transforms a broad class of smooth nonlinearities into polynomial surrogates, enabling perturbative continuation and one-shot transfer learning.
Concretely, we:
\begin{itemize}
  \item construct Chebyshev surrogates for general weak nonlinear terms, including non-polynomial response functions, over prescribed solution ranges;
  \item derive a perturbative reduction into a sequence of linear subproblems, each solved via one-shot output-layer updates using a pretrained multi-head PINN body;
  \item validate the proposed framework on nonlinear ODE and PDE benchmarks, demonstrating accuracy and fast online adaptation.
\end{itemize}

The proposed method targets many-query regimes in which the dominant linear operator is fixed and nonlinear effects are moderate over bounded solution ranges.

The remainder of the paper is organized as follows:
Section~\ref{sec:setting} introduces the problem setting,
Section~\ref{sec:method} presents the Chebyshev-augmented one-shot framework,
Section~\ref{sec:experiments} reports experimental results on nonlinear ODE and PDE benchmarks, Section~\ref{sec:discussion} concludes with a discussion of limitations and future directions, and Section~\ref{sec:conclusion} concludes the work.

\section{Related work}\label{sec:related}

\paragraph{PINNs and failure modes.}
PINNs trace back to early neural-network solvers for differential equations \citep{lagaris1998anns} and have been popularized by modern autograd-based formulations \citep{raissi2019pinns,karniadakis2021piml}. Known failure modes include spectral bias and poor extrapolation \citep{krishnapriyan2021failure,xu2019fprinciple}, motivating improved architectures, sampling, and training strategies.

\paragraph{Multi-instance training and transfer.}
Bundle-style and multi-head training amortize computation by training on multiple instances simultaneously \citep{flamant2020bundles,zou2023lhydra}. One-shot transfer learning for linear families provides closed-form adaptation by reducing inference to a least-squares solve in the output layer \citep{desai2022oneshot}. Operator-learning methods such as DeepONet and Fourier/Neural Operators provide alternative amortization mechanisms, learning maps between function spaces \citep{lu2021deeponet,li2021fno,kovachki2023neuraloperator}. Unlike operator-learning approaches, which aim to learn mappings between function spaces, one-shot transfer methods retain explicit equation structure and target fast adaptation under a fixed operator.

\paragraph{Perturbative PINNs for nonlinear problems.}
Perturbative OTL frameworks recover closed-form adaptation for certain nonlinear ODEs/PDEs by expanding solutions in a small parameter and solving a sequence of linear problems \citep{lei2023nonlinearodes,auroy2025nonlinearpdes,alexandrino2026ptl}. Our method retains the one-shot adaptation principle but uses Chebyshev approximation to broaden the nonlinearity class beyond polynomials.

\paragraph{Chebyshev approximation and spectral methods.}
Chebyshev polynomials provide near-minimax polynomial approximations with strong error guarantees for smooth analytic functions and form the backbone of many spectral methods \citep{trefethen2019atapia,boyd2001cheb}. We leverage these approximation properties not for discretization, but to construct polynomial surrogates that make nonlinear one-shot transfer tractable.

\section{Problem setting}\label{sec:setting}

We study one-shot transfer for families of weak nonlinear differential equations in a perturbative regime, targeting many-query settings. Here, a query refers to solving the same differential operator under new forcing terms, boundary conditions, or parameters.
Let $s$ denote the independent variables:
$s=t$ for ODEs and $s=(x,t)$ for PDEs, where $x\in\Omega_x\subset\mathbb{R}^d$
and $t\in[0,T]$.
We write $\Omega=\Omega_x\times[0,T]$ for space--time (with $\Omega=[0,T]$ in the ODE case)
and $\partial\Omega$ for the collection of boundary/initial sets.
For clarity we present a scalar unknown $u:\Omega\to\mathbb{R}$; extensions to vector-valued systems follow component-wise without conceptual changes.

We consider equation families of the form
\begin{equation}
\mathcal{D}u(s) + \varepsilon\,\mathcal{N}\!\big(u(s)\big) = f(s;\eta),
\qquad s\in\Omega,
\label{eq:setting_main_eps}
\end{equation}
subject to initial/boundary constraints
\begin{equation}
\mathcal{B}u(s)=b(s;\eta),\qquad s\in\partial\Omega,
\label{eq:setting_bc_eps}
\end{equation}
where $\mathcal{D}$ is a
fixed linear differential operator that defines the dominant dynamics and is shared across the family,
$\mathcal{N}$ is a pointwise nonlinearity which may be non-polynomial,
$\mathcal{B}$ collects the boundary/initial operators
(Dirichlet/Neumann and initial conditions),
and $\eta$ indexes instance-dependent inputs such as forcing terms and boundary/initial data.

The scalar $\varepsilon$ plays the role of a perturbation strength parameter.
In some applications, $\varepsilon$ is a physical coefficient already present in the model.
When the original equation does not include such a multiplier,
we treat $\varepsilon$ as a formal homotopy parameter;
we solve the family
$\mathcal{D}u + \varepsilon\,\mathcal{N}(u)=f$
and evaluate the approximation.
Equivalently, when admissible, one may induce a small effective $\varepsilon$ by rescaling
the unknown and/or coefficients so that the nonlinear contribution is moderate over the solution regime of interest.

Our objective is to design a method with a clear offline/online split.
Offline, we train once to learn a latent representation associated with the linear operator
$\mathcal{D}$ from a bundle of linear tasks.
Online, for each new nonlinear instance \eqref{eq:setting_main_eps}--\eqref{eq:setting_bc_eps} (i.e., each new query),
we compute an approximation of $u$
without retraining the network body or modifying the learned latent representation.
The online computation should reduce to a small number of closed-form linear solves
in the output layer (one-shot transfer), rather than iterative gradient-based optimization.

\section{Method}\label{sec:method}

\subsection{Overview}\label{sec:method:overview}

As described above, our framework follows a clear offline/online split: we learn an operator-aware representation offline once, then solve new nonlinear instances online via a small number of closed-form output-layer updates.

In the offline stage,
we train a multi-head PINN on a bundle of linear tasks that share the same linear operator $\mathcal{D}$.
This learns a shared feature map $H(s)\in\mathbb{R}^h$ (the network body), which captures operator-dependent structure.
After training, we freeze $H$.
With $H$ fixed, any linear instance
\[
\mathcal{D}u=g,\qquad \mathcal{B}u=h
\]
can be solved by a convex least-squares solve over the output-layer weights.
The corresponding system matrix used in the fit depends only on $\mathcal{D}$ and $\mathcal{B}$,
so it can be assembled and factorized once and reused across all the linear tasks.

In the online stage, given a nonlinear instance
\eqref{eq:setting_main_eps}--\eqref{eq:setting_bc_eps} (i.e., a new query),
we first approximate the pointwise nonlinearity $\mathcal{N}(u)$ by a truncated Chebyshev expansion
over a prescribed solution range $u\in[u_{\min},u_{\max}]$.
This produces a surrogate map $\mathcal{N}_m(u)$ represented in the Chebyshev basis whose coefficients are computed by Gauss--Chebyshev quadrature and whose evaluation uses the standard three-term recurrence (details in Appendix~\ref{app:chebyshev_surrogate}).

We then construct a truncated perturbation expansion in $\varepsilon$,
$u(s;\varepsilon)\approx \sum_{j=0}^{p}\varepsilon^j u_j(s)$,
substitute it into the surrogate equation
$\mathcal{D}u+\varepsilon\mathcal{N}_m(u)=f$,
and match powers of $\varepsilon$.
This yields $p+1$ linear subproblems sharing the operator $\mathcal{D}$;
the right-hand side at order $j$ depends only on the previously computed lower-order terms $\{u_0,\ldots,u_{j-1}\}$.
We impose evenly-splitted boundary/initial constraints on every order, so that the truncated series satisfies the original boundary/initial constraints up to truncation error.
Each linear subproblem is then solved by one-shot adaptation using the frozen feature map $H$.

Section~\ref{sec:method:chebyshev} states the Chebyshev surrogate used in the online stage, and Section~\ref{sec:method:perturb} presents the resulting perturbative linear recursion. Section~\ref{sec:method:training} describes the multi-head pretraining procedure used to learn a reusable operator-aware feature map for $\mathcal{D}$, while Section~\ref{sec:method:oneshot} presents the closed-form one-shot update that computes output weights. Finally, Section~\ref{sec:method:online} summarizes the complete online pipeline for solving nonlinear instances by sequential one-shot solves across perturbation orders and reconstructing $u$ from the truncated series.

\subsection{Chebyshev surrogate}\label{sec:method:chebyshev}

The goal of this step is to replace a general nonlinear term by a polynomial surrogate on a bounded solution range, enabling perturbative decomposition and closed-form adaptation in later stages.

We approximate the pointwise nonlinearity $\mathcal{N}(u)$ by a truncated Chebyshev series on a prescribed range
$u\in[u_{\min},u_{\max}]$ by mapping $u$ affinely to $\xi\in[-1,1]$ and expanding in Chebyshev polynomials of the first kind. Concretely, we use the surrogate
\begin{equation}
\mathcal{N}(u)\approx \mathcal{N}_m(u):=\sum_{\ell=0}^{m} c_\ell\,T_\ell(\Phi(u)),
\label{eq:cheb_trunc}
\end{equation}
where $\Phi:[u_{\min},u_{\max}]\to[-1,1]$ is the standard affine map.
The coefficients $\{c_\ell\}$ are computed by weighted Chebyshev projections, approximated in practice by Gauss--Chebyshev quadrature; evaluation uses the standard three-term recurrence. Full details of the affine map, coefficient formulas, quadrature rules, and stable evaluation are provided in Appendix~\ref{app:chebyshev_surrogate}.

\subsection{Perturbative expansion and linear subproblem recursion}\label{sec:method:perturb}

We solve the surrogate problem obtained by replacing $\mathcal{N}$ with $\mathcal{N}_m$ in
\eqref{eq:setting_main_eps}:
\begin{equation}
\mathcal{D}u(s)+\varepsilon\,\mathcal{N}_m\!\big(u(s)\big)=f(s;\eta),
\qquad \mathcal{B}u=b(\cdot;\eta).
\label{eq:surrogate_problem}
\end{equation}
We seek a truncated series in the same perturbation parameter $\varepsilon$, which we treat as small in the regime of interest, of the form
\begin{equation}
u(s;\varepsilon)\approx \sum_{j=0}^{p}\varepsilon^j u_j(s).
\label{eq:series_ansatz}
\end{equation}

Substituting \eqref{eq:series_ansatz} into \eqref{eq:surrogate_problem} requires expanding
$\mathcal{N}_m(u(s;\varepsilon))$ in powers of $\varepsilon$.
To preserve numerical stability, we perform this expansion directly in the Chebyshev basis using a Chebyshev recurrence lifted to truncated $\varepsilon$-series; the explicit forcing construction is given in Appendix~\ref{app:perturb_and_recursion}.
Matching powers of $\varepsilon$ yields
a sequence of linear problems with shared operator $\mathcal{D}$:
\begin{align}
\mathcal{D}u_0(s) &= f(s;\eta),
\qquad
\mathcal{B}u_0=b_0(\cdot;\eta),
\label{eq:order0}\\
\mathcal{D}u_j(s) &= -\mathcal{G}_{j-1}(s),
\qquad
\mathcal{B}u_j=b_j(\cdot;\eta),
\qquad j=1,\dots,p.
\label{eq:orderj}
\end{align}
Here, the boundary/initial data $\{b_j\}$ are chosen so that the truncated series
$\sum_{j=0}^{p}\varepsilon^j u_j$ satisfies the original constraints up to truncation error,
for example by evenly splitting the constraints across orders.
Thus the nonlinear problem reduces to $p+1$ linear subproblems governed by the same operator
$\mathcal{D}$. Each linear solve in \eqref{eq:order0}--\eqref{eq:orderj} is carried out by
one-shot adaptation with the frozen feature map $H$
(Section~\ref{sec:method:oneshot}).

\subsection{Multi-head pretraining on linear bundles}\label{sec:method:training}

All subproblems in \eqref{eq:order0}--\eqref{eq:orderj} share the same dominant linear operator
$\mathcal{D}$, so we learn a reusable feature map only once.
This offline training cost is amortized across all subsequent linear solves.

We parameterize the solution with a shared network body $H_\theta(s)$ and $K$ linear heads. A clear illustration of the multi-head architecture is shown in Appendix~\ref{app:mh_details}  (Figure~\ref{fig:mhpinn}).
The shared body is intended to capture operator-specific structure common to all tasks, while the heads account for task-specific forcing and boundary data.
In settings where the linear part contains higher-order derivatives, we use the standard first-order reformulation with auxiliary variables (details and an example are given in Appendix~\ref{app:mh_details}).

In the PDE case, it is convenient to view the frozen feature map as a matrix
$\mathbf{H}_\theta(s)\in\mathbb{R}^{2\times h}$ whose rows correspond to the two components
of the first-order state.
Each head uses a single weight vector $W_k\in\mathbb{R}^{h}$ that is shared across the components of the state vector, ensuring a consistent linear combination of the feature map for both the primary and auxiliary variables.
\begin{equation}
\hat{\mathbf{u}}^{(k)}(s)=\mathbf{H}_\theta(s)\,W_k,
\qquad
\text{i.e.,}\quad
\hat u^{(k)}(s)=H_{\theta,u}(s)^\top W_k,\ \ 
\hat y^{(k)}(s)=H_{\theta,y}(s)^\top W_k,
\qquad k=1,\dots,K.
\label{eq:mh_model_vec}
\end{equation}

We train the shared parameters $\theta$ and all head weights $\{W_k\}_{k=1}^K$ by minimizing a weighted sum of physics-informed residual losses and constraint losses (and an optional data loss when manufactured solutions are available). The explicit task specification and loss definitions are provided in Appendix~\ref{app:mh_details}.

\subsection{One-shot head solve for any linear subproblem}\label{sec:method:oneshot}

With the feature map $\mathbf{H}(s)=\mathbf{H}_\theta(s)$ frozen, each linear subproblem
sharing $\mathcal{D}$ is solved by optimizing only the output weights, following the
one-shot transfer construction in \citep{desai2022oneshot} and its perturbative extensions
\citep{lei2023nonlinearodes,auroy2025nonlinearpdes}.
We approximate the state by
\[
\hat{\mathbf{u}}(s)=\mathbf{H}(s)W,
\]
in particular the primary field is $\hat u(s)=H_u(s)^\top W$.

Let $\{s_n\}_{n=1}^{N_r}\subset\Omega$ and $\{\bar s_n\}_{n=1}^{N_b}\subset\partial\Omega$
be the fixed interior and constraint sampling sets.
These sampling sets are shared across all linear subproblems and perturbation orders.
Let $\mathbf{A}_r$ denote the stacked matrix collecting $(\mathcal{D}\mathbf{H})(s_n)$ over interior points, and let $\mathbf{A}_b$ denote the stacked matrix collecting $(\mathcal{B}H_u)(\bar s_n)^\top$ over constraint points; explicit stacking and dimensions are given in Appendix~\ref{app:oneshot_solve}.
The corresponding targets are stacked as
\[
\mathbf{f}^\ast := \big[\mathbf{f}^\ast(s_n)\big]_{n=1}^{N_r},
\qquad
\mathbf{b}^\ast := \big[b^\ast(\bar s_n)\big]_{n=1}^{N_b}.
\]
These targets encode the forcing and constraint data for the specific linear instance being solved.
The one-shot head is obtained by minimizing the same quadratic objective used in training
(with $\theta$ fixed):
\begin{equation}
\min_W\;
\frac{w_{\mathrm{pde}}}{N_r}\big\|\mathbf{A}_r W-\mathbf{f}^\ast\big\|_2^2
+
\frac{w_{\mathrm{bc}}}{N_b}\big\|\mathbf{A}_b W-\mathbf{b}^\ast\big\|_2^2.
\label{eq:oneshot_objective_updated}
\end{equation}
This objective is convex in $W$ and admits a unique solution under standard full-rank conditions; thus we have the closed-form update
\begin{equation}
W^\ast=\mathbf{M}^{-1}\mathbf{q}^\ast,
\label{eq:Wstar_updated}
\end{equation}
with
\begin{equation}
\mathbf{M}=\frac{w_{\mathrm{pde}}}{N_r}\mathbf{A}_r^\top\mathbf{A}_r
+
\frac{w_{\mathrm{bc}}}{N_b}\mathbf{A}_b^\top\mathbf{A}_b,
\qquad
\mathbf{q}^\ast=\frac{w_{\mathrm{pde}}}{N_r}\mathbf{A}_r^\top\mathbf{f}^\ast
+
\frac{w_{\mathrm{bc}}}{N_b}\mathbf{A}_b^\top\mathbf{b}^\ast.
\label{eq:M_q_updated}
\end{equation}
For fixed $\mathcal{D}$, $\mathcal{B}$, frozen $\mathbf{H}$, and fixed sampling sets,
$\mathbf{M}$ is constant across instances and across perturbation orders.
Consequently, $\mathbf{M}$ depends only on the linear operator, the constraint type, and the sampling strategy.
Hence $\mathbf{M}^{-1}$ can be precomputed once and reused,
and each new linear solve reduces to constructing $\mathbf{q}^\ast$.
This reuse is the key to achieving fast online adaptation in many-query settings.

\subsection{Online solve for nonlinear instances}\label{sec:method:online}

Given a nonlinear instance \eqref{eq:setting_main_eps}--\eqref{eq:setting_bc_eps},
the online stage performs a sequence of one-shot linear solves while keeping the
pretrained feature map $\mathbf{H}$ fixed.
No gradient-based retraining is performed in the online stage.
We first select a working range $[u_{\min},u_{\max}]$ and construct the Chebyshev surrogate
$\mathcal{N}_m$ as in Section~\ref{sec:method:chebyshev}.
We then compute the perturbation coefficients $\{u_j\}_{j=0}^{p}$ by solving the linear
subproblems \eqref{eq:order0}--\eqref{eq:orderj} sequentially.

For each order $j$, the linear subproblem provides an interior forcing $g_j(s)$
(with $g_0=f$ and, for $j\ge 1$, $g_j$ determined by the Chebyshev-based forcing construction
in Appendix~\ref{app:perturb_and_recursion})
and a constraint target $b_j$ consistent with the constraint splitting used in
\eqref{eq:order0}--\eqref{eq:orderj}.
Both the forcing and constraint targets are assembled deterministically from previously computed lower-order solutions.
In the first-order PDE example (Appendix~\ref{app:mh_details}),
we lift the scalar forcing to a vector target by
\[
\mathbf{f}^{(j)}(s)=
\begin{bmatrix}
g_j(s)\\
0
\end{bmatrix},
\]
and apply the constraint operator only to the primary component $u_j$. With the fixed matrices $\mathbf{A}_r,\mathbf{A}_b$ (Appendix~\ref{app:oneshot_solve}),
we form the corresponding stacked targets
$\mathbf{f}^{(j)}$ and
$\mathbf{b}^{(j)}$,
assemble $\mathbf{q}^{(j)}$ from \eqref{eq:M_q_updated},
and compute the head in closed form:
\begin{equation}
W_j=\mathbf{M}^{-1}\mathbf{q}^{(j)},
\qquad j=0,1,\dots,p.
\label{eq:online_Wj_updated}
\end{equation}
Since $\mathbf{M}^{-1}$ is reused, the cost of each order-$j$ solve is dominated by forming the right-hand side $\mathbf{q}^{(j)}$.
The order-$j$ solutions are then given by the primary-component reconstruction
$u_j(s)=H_u(s)^\top W_j$,
and the final approximation is
\begin{equation}
u(s;\varepsilon)
=
\sum_{j=0}^{p}\varepsilon^j u_j(s)
=
\sum_{j=0}^{p}\varepsilon^j H_u(s)^\top W_j.
\label{eq:final_reconstruction_updated}
\end{equation}
This reconstruction completes the online solve for a nonlinear query using only closed-form linear algebra operations.

\subsection{Algorithmic summary}\label{sec:method:algorithm}

An algorithmic summary of the offline/online pipeline is provided in Appendix~\ref{app:alg_details}.

\section{Results}\label{sec:experiments}

We evaluate the proposed Chebyshev-augmented one-shot perturbative PINN on three nonlinear benchmarks:
two second-order ODEs with pointwise nonlinearities and one reaction--diffusion PDE with a rational reaction term.
The goal of these experiments is to assess accuracy and online efficiency under one-shot nonlinear transfer. And to demonstrate the efficiency, we compare the runtime of our method with a regular baseline where we freeze the pretrained feature map and retrains only the linear heads via gradient descent at test time.
All experiments follow the offline/online workflow described in Section~\ref{sec:method}. We report mean squared error (MSE) metrics and time per online solve; full implementation details, architectures, sampling, metric definitions, timing protocol, baseline training settings, and ablation results are provided in Appendix~\ref{app:exp_details} and Appendix~\ref{app:ablations}.

\subsection{Benchmarks and evaluation setup}\label{sec:experiments:setup}

\paragraph{ODE benchmarks.}
We consider the following families on $t\in[0,5]$:
\begin{align}
\textbf{ODE1:}\quad
u''(t)+\delta\,u'(t)+\alpha\,u(t)+\varepsilon\cos\!\big(u(t)\big) &=\beta\cos(\omega t),
\label{eq:exp_ode1}\\
\textbf{ODE2:}\quad
u''(t)+\delta\,u'(t)+\alpha\,u(t)+\varepsilon\,u(t)^{-2} &=\gamma-e^{-t}.
\label{eq:exp_ode2}
\end{align}
ODE1 tests the ability of the surrogate-and-perturbation pipeline to handle a smooth non-polynomial nonlinearity $\cos(u)$,
which cannot be represented exactly by finite polynomial perturbations,
while ODE2 probes a more challenging singular nonlinearity $u^{-2}$ (the test distribution is chosen to keep trajectories away from $u=0$; see Appendix~\ref{app:exp_details}).

\paragraph{PDE benchmark.}
On $\Omega=[0,1]\times[0,1]$, we consider
\begin{equation}
\textbf{PDE1:}\quad
u_t(x,t)=D\,u_{xx}(x,t)+\varepsilon\left(\frac{u(x,t)}{u(x,t)+1}-\delta\,u(x,t)\right)+f(x,t),
\label{eq:exp_pde1}
\end{equation}
with constant boundary/initial conditions and a forcing term $f$ derived from manufactured solutions. This benchmark targets a reaction--diffusion regime where the dominant linear operator is parabolic and the nonlinear reaction term is rational in $u$.

\paragraph{Baseline: head retraining.}
In addition to our closed-form one-shot online solve, we consider a baseline that preserves the same pretrained feature map but replaces the closed-form head computation with iterative optimization.
Specifically, we freeze the shared body $H_\theta$ learned in the offline stage and optimize only the linear head $W$ for the target nonlinear instance using gradient descent on the same objective function.
Training hyperparameters and settings are reported in Appendix~\ref{app:head_gd_details}.

\paragraph{What is measured.}
For ODE1--ODE2, we report the mean squared equation residual evaluated on a uniform grid of $100$ time points and averaged across $100$ test instances. For PDE1, we report the mean squared error against the manufactured truth on a fixed $61\times 61$ evaluation grid, averaged across $32$ test instances. All reported online times exclude offline training and the one-time precomputation of $\mathbf{M}^{-1}$ and reflect only the per-query cost of the online stage
(Appendix~\ref{app:timing_details}). For the baseline, we report the wall-clock time to reach a prescribed error threshold (early stopping).
Specifically, we stop learning once the mean squared error falls below $\tau\in\{5\times 10^{-4},\,5\times 10^{-3},\,10^{-2}\}$ respectively for each benchmark and record the average time of the runs that successfully reach the threshold.

\subsection{Summary}\label{sec:experiments:summary}

Table~\ref{tab:exp_summary} summarizes the main quantitative outcomes across all three benchmarks, including the online settings used (nonlinearity strength $\varepsilon$, perturbation order $p$, Chebyshev degree $m$, and quadrature size $M$). Across all cases, the online stage consists of sequential one-shot solves across orders $\{u_j\}_{j=0}^{p}$ with a fixed pretrained feature map and a task-invariant inverse $\mathbf{M}^{-1}$.

\begin{table}[H]
\caption{Summary results of our method on all the systems investigated. Training with a small number of heads is sufficient to enable accurate one-shot transfer to unseen nonlinear instances within the prescribed regime. All times are reported for runs on a CPU.}
\label{tab:exp_summary}
\begin{center}
\small
\begin{tabular}{lccccc}
\multicolumn{1}{c}{\bf Benchmark} &
\multicolumn{1}{c}{\bf Nonlinearity} &
\multicolumn{1}{c}{\bf Online $(\varepsilon,p,m,M)$} &
\multicolumn{1}{c}{\bf MSE} &
\multicolumn{1}{c}{\bf Online time (s)} &
\multicolumn{1}{c}{\bf Baseline time (s)}
\\ \hline \\
ODE1 & $\cos(u)$ & $(0.5,\,12,\,20,\,1000)$ & $3.83\times 10^{-6}$ & $9.63\times 10^{-2}$ & $8.86$ \\
ODE2 & $u^{-2}$ & $(0.1,\,12,\,20,\,1000)$ & $5.31\times 10^{-5}$ & $7.85\times 10^{-2}$ & $24.95$\\
PDE1 & $\frac{u}{u+1}-\delta u$ & $(0.5,\,20,\,30,\,1000)$ & $7.12\times 10^{-5}$ & $9.61\times 10^{-2}$ & $221.58$\\
\hline
\end{tabular}
\end{center}
\end{table}

\subsection{ODE1 results: cosine nonlinearity}\label{sec:experiments:ode1}

\begin{figure}[H]
\centering
\includegraphics[width=0.85\textwidth]{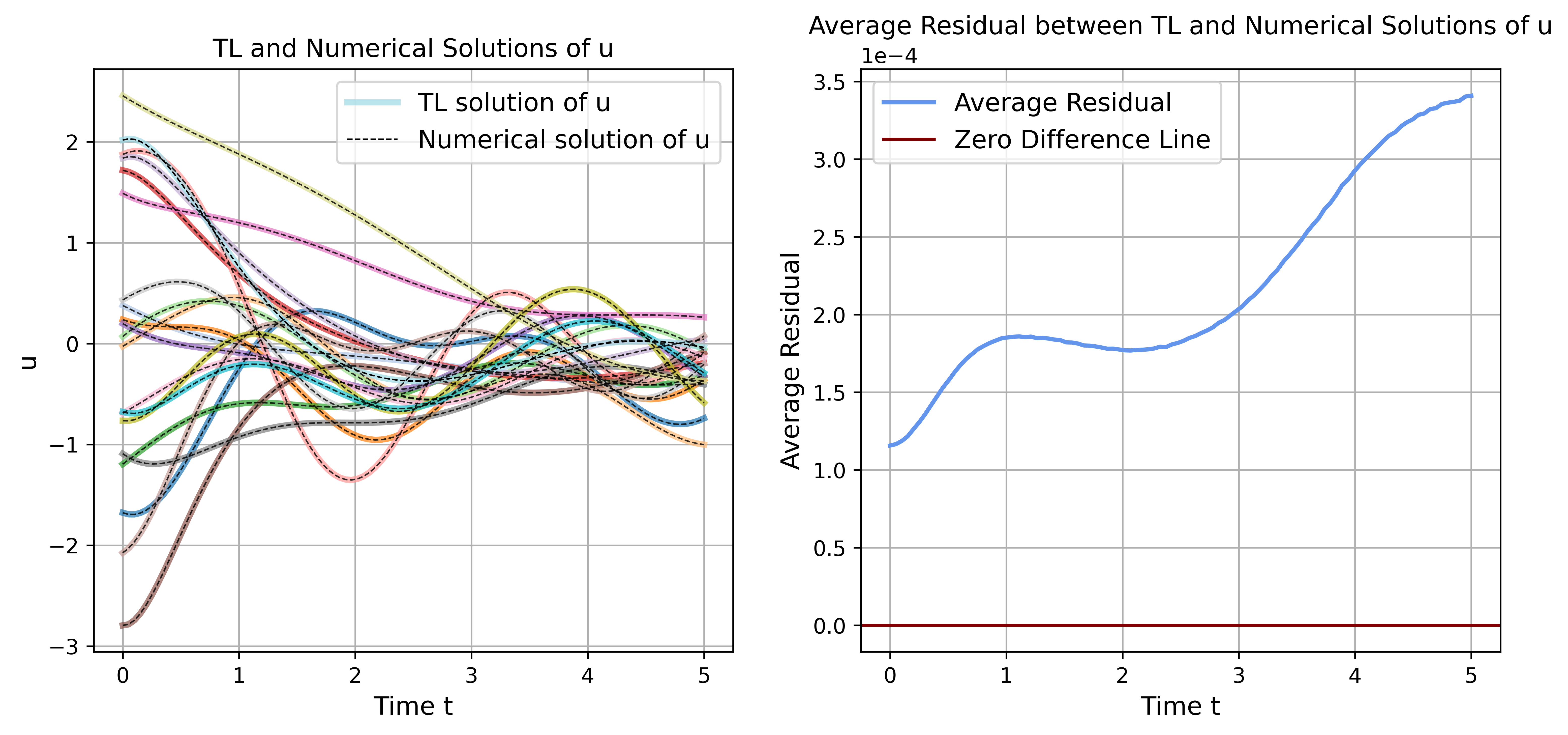}
\caption{ODE1 \eqref{eq:exp_ode1}. \textbf{Left:} overlay of one-shot transfer predictions $u_{\mathrm{TL}}(t)$ (solid) and numerical reference solutions $u_{\mathrm{ref}}(t)$ (dashed) for a representative subset of test instances. \textbf{Right:} mean solution discrepancy over the same subset of test instances (red line indicates $\Delta(t)=0$).}
\label{fig:ode1}
\end{figure}

Using the configuration in Table~\ref{tab:exp_summary}, ODE1 achieves a mean equation-residual MSE of $3.83\times 10^{-6}$ across $100$ test instances, while retaining $9.63\times 10^{-2}$s online time per instance. Beyond residual metrics, we compare the reconstructed series solution against a numerical reference solver (RK45). As shown in Figure~\ref{fig:ode1} (left), the one-shot transfer prediction closely tracks the numerical trajectories
despite being obtained without any gradient-based retraining at test time
for a subset of test instances.

To visualize the agreement over time, Figure~\ref{fig:ode1} (right) plots the mean discrepancy
\[
\Delta(t)\;:=\;\frac{1}{N}\sum_{i=1}^{N}\Big(u_{\mathrm{TL}}^{(i)}(t)-u_{\mathrm{ref}}^{(i)}(t)\Big),
\]
computed over the test suite on the same uniform grid. The discrepancy remains small throughout $[0,5]$, indicating that the online surrogate-and-perturbation recursion yields a stable reconstruction over the time interval.

\subsection{ODE2 results: inverse-square nonlinearity}\label{sec:experiments:ode2}

\begin{figure}[H]
\centering
\includegraphics[width=0.85\textwidth]{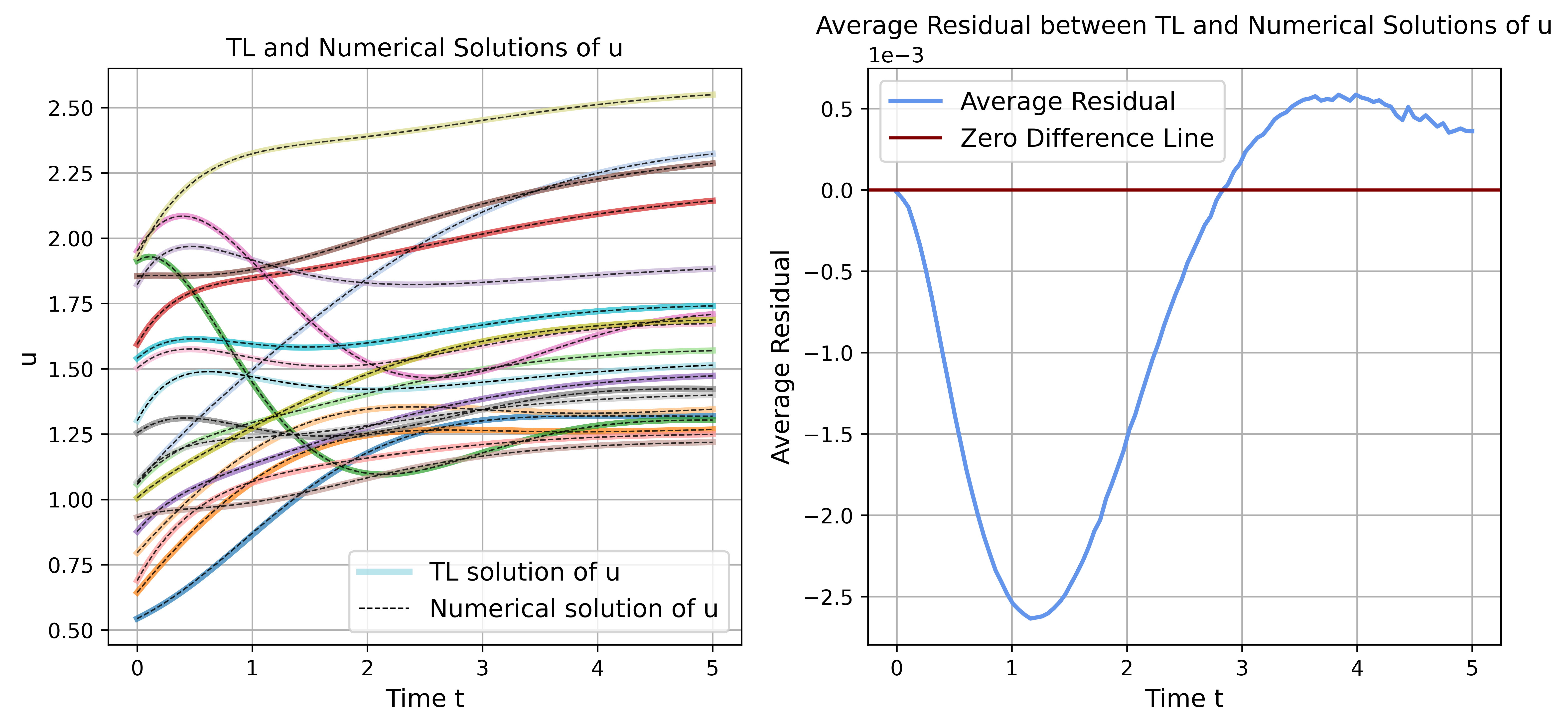}
\caption{ODE2 \eqref{eq:exp_ode2}. \textbf{Left:} overlay of one-shot transfer predictions $u_{\mathrm{TL}}(t)$ (solid) and numerical reference solutions $u_{\mathrm{ref}}(t)$ (dashed) for a representative subset of test instances. \textbf{Right:} mean solution discrepancy over the same subset of test instances (red line indicates $\Delta(t)=0$).}
\label{fig:ode2}
\end{figure}

ODE2 is more sensitive due to the inverse-square term $u^{-2}$, which can amplify errors if trajectories approach $u=0$. Under the configuration in Table~\ref{tab:exp_summary}, the method attains a mean equation-residual MSE of $5.31\times 10^{-5}$ across $100$ instances with an average online time of $7.85\times 10^{-2}$ seconds per solve. Despite the increased difficulty,
and the potential amplification of approximation errors near singularities,
the trajectory overlays in Figure~\ref{fig:ode2} (left) show that the one-shot transfer solutions remain in close agreement with the numerical references across the interval.

Figure~\ref{fig:ode2} (right) reports the mean discrepancy $\Delta(t)$ as in ODE1. The discrepancy remains small in magnitude relative to the solution scale in Figure~\ref{fig:ode2} (left), supporting the feasibility of the Chebyshev surrogate combined with one-shot perturbative solves in this regime.

\subsection{PDE1 results: reaction--diffusion with manufactured forcing}\label{sec:experiments:pde1}

For PDE1, the forcing term $f(x,t)$ is constructed from manufactured solutions, enabling direct evaluation of solution error on a fixed grid. Using the configuration in Table~\ref{tab:exp_summary}, the method achieves an average solution MSE of $7.12\times 10^{-5}$ on a $61\times 61$ evaluation grid, with an average online time of $9.61\times 10^{-2}$ seconds per solve.

\begin{figure}[H]
\centering
\includegraphics[width=0.85\textwidth]{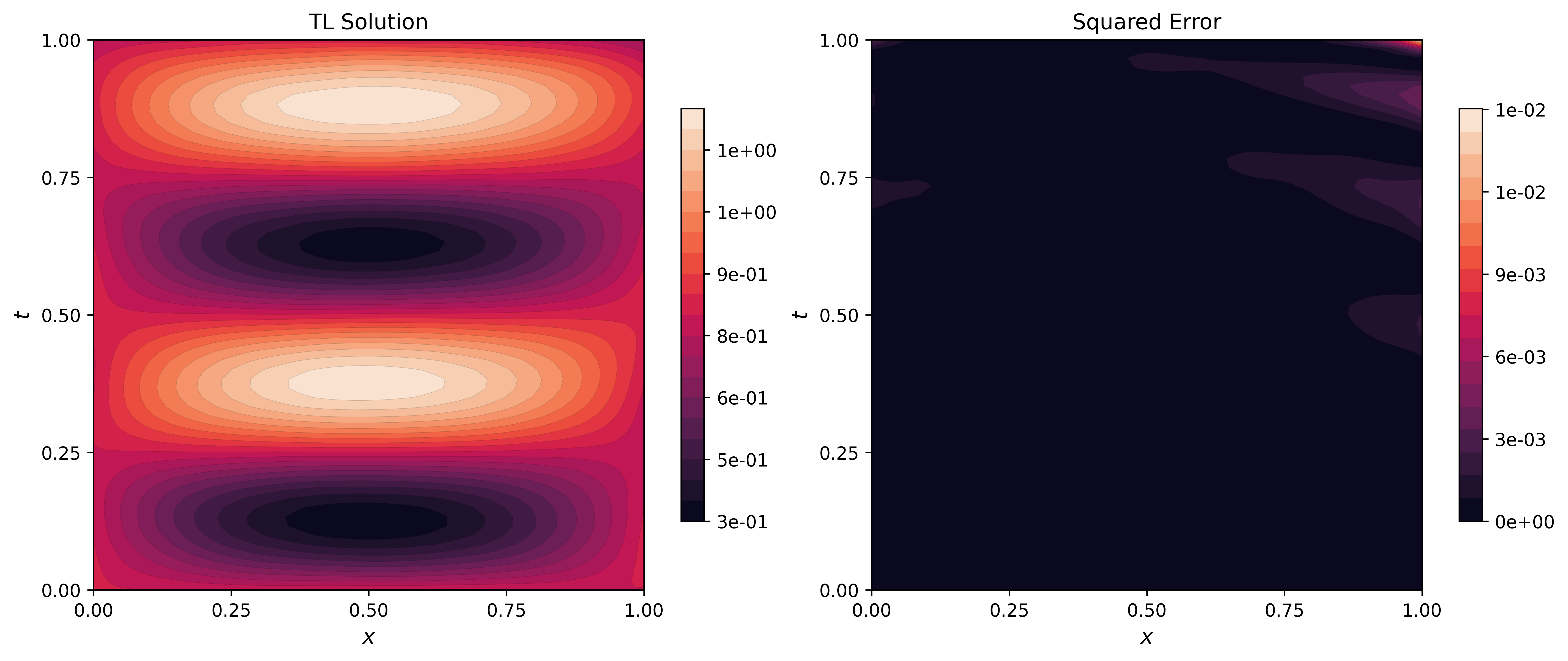}
\caption{PDE1 \eqref{eq:exp_pde1}. \textbf{Left:} predicted field $u_{\mathrm{TL}}(x,t)$ on the $[0,1]\times[0,1]$ space--time domain. \textbf{Right:} pointwise squared error $\big(u_{\mathrm{TL}}(x,t)-u_{\mathrm{true}}(x,t)\big)^2$ against the manufactured solution on the same grid.}
\label{fig:pde1}
\end{figure}

Figure~\ref{fig:pde1} visualizes a representative prediction. The left panel shows the predicted field $u_{\mathrm{TL}}(x,t)$, while the right panel reports the pointwise squared error $\big(u_{\mathrm{TL}}(x,t)-u_{\mathrm{true}}(x,t)\big)^2$. The error remains small over most of the domain,
with localized increases near the boundary regions in this instance.
These boundary-localized errors are consistent with truncation effects in the perturbative reconstruction and finite Chebyshev approximation.

\subsection{Brief discussion}\label{sec:experiments:discussion}

Across all three benchmarks, the proposed method achieves low MSE with relatively short online solve times per instance. ODE1 demonstrates strong agreement for a smooth bounded nonlinearity, while ODE2 highlights that the approach remains effective for a singular inverse-power term under a small-nonlinearity regime. PDE1 confirms that the same offline/online pipeline transfers to a spatiotemporal parabolic operator
with a rational reaction term, yielding accurate reconstructions against a manufactured ground truth.

In addition, we notice that the conventional gradient descent baseline is substantially slower at test time even though it freezes the same pretrained feature map and optimizes only the final linear head. This is orders of magnitude larger than the $\mathcal{O}(10^{-1})$s per-instance online cost of our one-shot solves in Table~\ref{tab:exp_summary}. Moreover, the baseline is less reliable across instances and can fail to reach comparable accuracy without many iterations, highlighting both the efficiency and robustness benefits of the proposed online adaptation framework.

Taken together, these results demonstrate that Chebyshev surrogates can effectively extend one-shot perturbative PINNs beyond polynomial nonlinearities while retaining fast online adaptation.
To quantify sensitivity to key online hyperparameters, we perform ablation studies over perturbative strength $\varepsilon$, perturbation order $p$, and Chebyshev degree $m$ for ODE1 and PDE1 in Appendix~\ref{app:ablations}.
We also report training and optimization settings for the baseline in Appendix~\ref{app:head_gd_details}.

\section{Discussion}\label{sec:discussion}

Our framework inherits the strengths of one-shot transfer for linear operators:
fast inference and reusability of latent features \citep{desai2022oneshot}.
Meanwhile, it extends applicability to a broad class of nonlinearities via Chebyshev surrogates.
This combination enables efficient many-query adaptation while preserving explicit equation structure.

\paragraph{Limitations.}
The method is not a universal replacement for operator learning.
When nonlinearities are very strong,
solutions leave bounded ranges,
or chaotic behaviors
develop,
surrogate polynomials may require high degree or piecewise treatment,
and the perturbation sequence may become unstable.
In such regimes, full operator-learning approaches or instance-specific PINN training strategies may be more appropriate.

% ============================================================
\section{Conclusion}\label{sec:conclusion}

We presented a Chebyshev-augmented framework for one-shot transfer learning in PINNs
applied to nonlinear ODEs and PDEs.
By approximating general nonlinear terms with Chebyshev polynomials and coupling this
surrogate with a perturbative expansion,
we reduce a nonlinear problem to a sequence of linear subproblems,
each solvable by closed-form output-layer updates in a pretrained latent space.
This construction preserves the efficiency and reusability of one-shot transfer while extending its applicability beyond polynomial nonlinearities.

The resulting method targets many-query settings,
offering fast adaptation across new forcing terms and boundary/initial conditions
without retraining network bodies or modifying the learned feature representation.
This framework is particularly relevant for practitioners
who seek to rapidly obtain approximate solutions to differential equations
with the same dominant operator
but under varying initial/boundary conditions, forcings, or coefficients.

To our knowledge, this is the first framework that combines Chebyshev approximation with one-shot perturbative PINNs to enable closed-form nonlinear adaptation beyond polynomial nonlinearities.

Future directions include adaptive or piecewise Chebyshev surrogates to extend the perturbative regime, integration with data-rich or high-dimensional settings, and further optimization of the online pipeline for large-scale many-query applications.

\subsection*{Declaration of AI usage}\label{sec:AI}
The authors used ChatGPT to assist with language refinement to improve readability and wording. The tool was not used to generate experimental results or to formulate mathematical arguments. All equations, derivations, and interpretations were critically reviewed by the authors, and the authors take full responsibility for the contents of the work.

\bibliography{iclr2026_conference}
\bibliographystyle{iclr2026_conference}

\appendix
\section{Appendix}

The code has been made publicly available on: \url{https://github.com/ryqherry/Cheby-PINNs}.

\subsection{Additional method details}\label{app:method_details}

\subsubsection{Chebyshev surrogate construction via Gauss--Chebyshev quadrature}\label{app:chebyshev_surrogate}

The goal of this step is to replace a general nonlinear term by a polynomial surrogate on a bounded solution range, enabling perturbative decomposition and closed-form adaptation in later stages.

We approximate the pointwise nonlinearity $\mathcal{N}(u)$ by a truncated Chebyshev series on a prescribed range
$u\in[u_{\min},u_{\max}]$. Define the affine map $\Phi:[u_{\min},u_{\max}]\to[-1,1]$ by
\begin{equation}
\xi = \Phi(u) := \frac{2u-(u_{\max}+u_{\min})}{u_{\max}-u_{\min}},
\qquad
u = \Phi^{-1}(\xi)=\frac{u_{\max}-u_{\min}}{2}\xi+\frac{u_{\max}+u_{\min}}{2}.
\label{eq:cheb_affine_map}
\end{equation}
Let $\tilde{\mathcal{N}}(\xi):=\mathcal{N}(\Phi^{-1}(\xi))$.

We employ Chebyshev polynomials of the first kind $\{T_\ell\}_{\ell\ge 0}$ defined by the three-term recurrence
\begin{equation}
T_0(\xi)=1,\qquad T_1(\xi)=\xi,\qquad T_{\ell+1}(\xi)=2\xi\,T_\ell(\xi)-T_{\ell-1}(\xi),\quad \ell\ge 1.
\label{eq:cheb_recurrence}
\end{equation}
In the main paper we use the truncated surrogate
$\mathcal{N}(u)\approx \mathcal{N}_m(u)$ defined in \eqref{eq:cheb_trunc}.

\noindent The coefficients are given by weighted Chebyshev projections:
\begin{align}
c_0
&=\frac{1}{\pi}\int_{-1}^{1}\frac{\tilde{\mathcal{N}}(\xi)}{\sqrt{1-\xi^2}}\,d\xi,
\label{eq:c0_int}\\
c_\ell
&=\frac{2}{\pi}\int_{-1}^{1}\frac{\tilde{\mathcal{N}}(\xi)\,T_\ell(\xi)}{\sqrt{1-\xi^2}}\,d\xi,
\qquad \ell\ge 1.
\label{eq:cl_int}
\end{align}
We approximate \eqref{eq:c0_int}--\eqref{eq:cl_int} using Gauss--Chebyshev quadrature (first kind). With $M$ quadrature nodes $\xi_j=\cos\theta_j$ and $\theta_j=\frac{(2j-1)\pi}{2M}$, $j=1,\dots,M$, we use
\begin{align}
c_0 &\approx \frac{1}{M}\sum_{j=1}^{M}\tilde{\mathcal{N}}(\xi_j),
\label{eq:c0_gc}\\
c_\ell &\approx \frac{2}{M}\sum_{j=1}^{M}\tilde{\mathcal{N}}(\xi_j)\,T_\ell(\xi_j),
\qquad \ell\ge 1.
\label{eq:cl_gc}
\end{align}
In implementation, for each node $\xi_j$ we compute $\{T_\ell(\xi_j)\}_{\ell=0}^{m}$ by iterating \eqref{eq:cheb_recurrence}. We choose $M$ sufficiently large to reduce numerical error when $\tilde{\mathcal{N}}$ varies rapidly near $\xi=\pm 1$.

Given coefficients $\{c_\ell\}$, evaluation of $\mathcal{N}_m(u)$ at a point proceeds by computing $\xi=\Phi(u)$, generating $\{T_\ell(\xi)\}_{\ell=0}^{m}$ via \eqref{eq:cheb_recurrence}, and forming the sum in \eqref{eq:cheb_trunc}. We keep the Chebyshev representation throughout; in particular, we do not require any conversion to a monomial basis to carry out the perturbative recursion below.

\subsubsection{Perturbative expansion and linear subproblem recursion}\label{app:perturb_and_recursion}

Substituting the truncated series ansatz \eqref{eq:series_ansatz} into the surrogate problem \eqref{eq:surrogate_problem} requires expanding $\mathcal{N}_m(u(s;\varepsilon))$ in powers of $\varepsilon$.
To preserve numerical stability, we perform this expansion directly in the Chebyshev basis.

Let $\xi(s;\varepsilon):=\Phi(u(s;\varepsilon))$.
Since $\Phi$ is affine,
\begin{equation}
\xi(s;\varepsilon)=\xi_0(s)+\sum_{j=1}^{p}\varepsilon^j \xi_j(s),
\qquad
\xi_0(s)=\Phi(u_0(s)),
\qquad
\xi_j(s)=\alpha\,u_j(s)\ (j\ge 1),
\label{eq:xi_series}
\end{equation}
where $\alpha=\frac{2}{u_{\max}-u_{\min}}$.

\vspace{0.3cm} 
\noindent For each $\ell$, define coefficients $\tau_{\ell,j}(s)$ by the truncated series
\begin{equation}
T_\ell\!\big(\xi(s;\varepsilon)\big)
=
\sum_{j=0}^{p}\varepsilon^j\,\tau_{\ell,j}(s)
+\mathcal{O}(\varepsilon^{p+1}).
\label{eq:tau_def}
\end{equation}
That is, $\tau_{\ell,j}(s)$ denotes the coefficient of $\varepsilon^j$ in the expansion of
$T_\ell(\xi(s;\varepsilon))$.
We compute $\{\tau_{\ell,j}\}$ using the Chebyshev recurrence lifted to $\varepsilon$-series.
Initialize
\begin{equation}
\tau_{0,0}=1,\quad \tau_{0,j}=0\ (j\ge 1),
\qquad
\tau_{1,j}=\xi_j,\quad j=0,1,\dots,p,
\label{eq:tau_init}
\end{equation}
and for $\ell\ge 1$ define
\begin{equation}
\tau_{\ell+1,j}
=
2\sum_{k=0}^{j}\xi_k\,\tau_{\ell,j-k}
-
\tau_{\ell-1,j},
\qquad j=0,1,\dots,p,
\label{eq:tau_rec}
\end{equation}
where the sum $\sum_{k=0}^{j}\xi_k\,\tau_{\ell,j-k}$ is the coefficient of $\varepsilon^{j}$
in the product $\xi(s;\varepsilon)\,T_\ell(\xi(s;\varepsilon))$ for the recurrence.
This follows from the Cauchy product of truncated $\varepsilon$-series.
That is, multiplying the truncated $\varepsilon$-series
$\xi(s;\varepsilon)=\sum_{k=0}^{p}\varepsilon^k\xi_k(s)$ and
$T_\ell(\xi(s;\varepsilon))=\sum_{r=0}^{p}\varepsilon^r\tau_{\ell,r}(s)$
yields the coefficient
\[
[\varepsilon^{j}]\,\xi\,T_\ell
=
\sum_{k=0}^{j}\xi_k\,\tau_{\ell,j-k}.
\]

\noindent Using \eqref{eq:cheb_trunc} and \eqref{eq:tau_def}, the surrogate nonlinearity expands as
\begin{equation}
\mathcal{N}_m(u(s;\varepsilon))
=
\sum_{\ell=0}^{m} c_\ell\,T_\ell\!\big(\xi(s;\varepsilon)\big)
=
\sum_{j=0}^{p}\varepsilon^j\,\mathcal{G}_j(s)
+\mathcal{O}(\varepsilon^{p+1}),
\qquad
\mathcal{G}_j(s):=\sum_{\ell=0}^{m} c_\ell\,\tau_{\ell,j}(s).
\label{eq:Gj_def}
\end{equation}
Thus, $\mathcal{G}_j(s)$ collects all contributions to the nonlinearity at order $\varepsilon^j$ and depends only on lower-order solution components through $\tau_{\ell,j}$.

Substituting \eqref{eq:series_ansatz} and \eqref{eq:Gj_def} into \eqref{eq:surrogate_problem} and matching powers of $\varepsilon$ yields the linear subproblems \eqref{eq:order0}--\eqref{eq:orderj} in the main paper.

\subsubsection{Multi-head pretraining details}\label{app:mh_details}

All subproblems in \eqref{eq:order0}--\eqref{eq:orderj} share the same dominant linear operator
$\mathcal{D}$, so we learn a reusable feature map only once.
This offline training cost is amortized across all subsequent linear solves.
For differential equations whose linear part contains higher-order derivatives,
our implementation follows the standard first-order reformulation:
we introduce auxiliary variables so that the linear operator acts on a vector-valued state.
For example, for a diffusion-type operator $u_t-\kappa u_{xx}$ we introduce $y:=u_x$
and define the state $\mathbf{u}:=[u,y]^\top$, yielding the linear system
\begin{equation}
\mathcal{D}\mathbf{u}=
\begin{bmatrix}
u_t-\kappa\,y_x\\ u_x-y
\end{bmatrix},
\qquad
\mathbf{f}(s)=\begin{bmatrix} f(s)\\ 0\end{bmatrix},
\label{eq:first_order_system}
\end{equation}
with constraints applied to the primary component $u$
(e.g.\ Dirichlet/initial values).
In ODE settings, similar auxiliary variables can be introduced when
$\mathcal{D}$ contains higher-order derivatives \citep{lei2023nonlinearodes}.

We parameterize the solution with a shared network body $H_\theta(s)$ and $K$ linear heads as in \eqref{eq:mh_model_vec}. The following Figure~\ref{fig:mhpinn} illustrates the general architecture of the multi-head PINN we use for pretraining

\begin{figure}[H]
\centering
\includegraphics[width=0.85\textwidth]{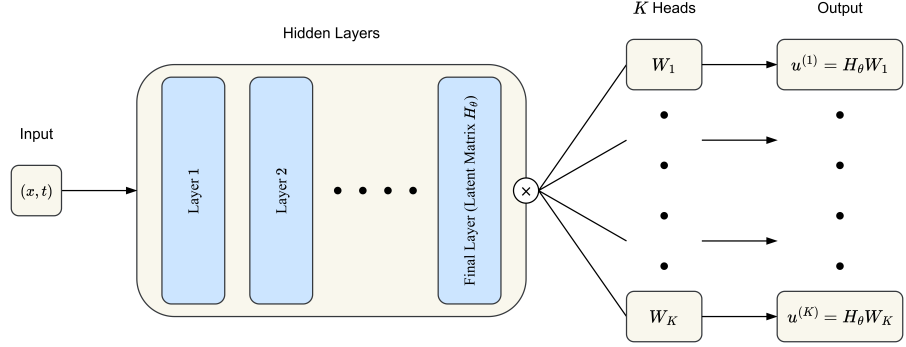}
\caption{Multi-head PINN architecture used in the offline stage. A shared body produces features $H_\theta(s)$ and each head corresponds to a linear output layer with weights $W_k$, $k=1,\dots,K$.}
\label{fig:mhpinn}
\end{figure}

For head $k$ we specify a linear instance with fixed $(\mathcal{D},\mathcal{B})$ and
task-dependent targets
\begin{equation}
\mathcal{D}\mathbf{u}=\mathbf{f}_k \ \text{in }\Omega,
\qquad
\mathcal{B}u=b_k \ \text{on }\partial\Omega,
\label{eq:train_task_k_vec}
\end{equation}
where $\mathcal{B}$ acts on the primary field $u$
(the first component of $\mathbf{u}$).
The forcing $\mathbf{f}_k$ and boundary/initial data $b_k$ vary across heads, while the operator $\mathcal{D}$ and constraint type $\mathcal{B}$ remain fixed.
Let $\{s_n\}_{n=1}^{N_r}\subset\Omega$ be interior points and
$\{\bar s_n\}_{n=1}^{N_b}\subset\partial\Omega$ be constraint points.We define the per-head physics-informed losses that enforce the governing equations and constraints for each linear task in the training bundle
\begin{align}
\mathcal{L}^{(k)}_{\mathrm{pde}}(\theta,W_k)
&:=
\frac{1}{N_r}\sum_{n=1}^{N_r}
\Big\|\mathcal{D}\hat{\mathbf{u}}^{(k)}(s_n)-\mathbf{f}_k(s_n)\Big\|_2^2,
\label{eq:train_loss_pde_vec}\\
\mathcal{L}^{(k)}_{\mathrm{bc}}(\theta,W_k)
&:=
\frac{1}{N_b}\sum_{n=1}^{N_b}
\Big|\mathcal{B}\hat u^{(k)}(\bar s_n)-b_k(\bar s_n)\Big|^2.
\label{eq:train_loss_bc_u}
\end{align}

\noindent When manufactured solutions are available for the training bundle,
we add a data loss term
\begin{equation}
\mathcal{L}^{(k)}_{\mathrm{data}}(\theta,W_k)
:=
\frac{1}{N_r}\sum_{n=1}^{N_r}
\Big|\hat u^{(k)}(s_n)-u_k^{\mathrm{ref}}(s_n)\Big|^2,
\label{eq:train_loss_data}
\end{equation}
and set its weight to zero otherwise.

We train the shared parameters $\theta$ and all head weights $\{W_k\}_{k=1}^K$ by minimizing
\begin{equation}
\min_{\theta,\{W_k\}}\;
\sum_{k=1}^{K}\left(
w_{\mathrm{pde}}\,\mathcal{L}^{(k)}_{\mathrm{pde}}(\theta,W_k)
+
w_{\mathrm{bc}}\,\mathcal{L}^{(k)}_{\mathrm{bc}}(\theta,W_k)
+
w_{\mathrm{data}}\,\mathcal{L}^{(k)}_{\mathrm{data}}(\theta,W_k)
\right),
\label{eq:mh_training_objective_updated}
\end{equation}
with $w_{\mathrm{pde}},w_{\mathrm{bc}}\ge 0$ and $w_{\mathrm{data}}\ge 0$.
After training, we freeze $\theta$ (hence $\mathbf{H}$) and discard the training heads; only the frozen feature map is retained for all subsequent one-shot solves on new tasks.

\subsubsection{One-shot head solve construction}\label{app:oneshot_solve}

Let $\{s_n\}_{n=1}^{N_r}\subset\Omega$ and $\{\bar s_n\}_{n=1}^{N_b}\subset\partial\Omega$
be the fixed interior and constraint sampling sets.
These sampling sets are shared across all linear subproblems and perturbation orders.
We form the stacked operator--feature matrices
\begin{equation}
\mathbf{A}_r :=
\begin{bmatrix}
(\mathcal{D}\mathbf{H})(s_1)\\
\vdots\\
(\mathcal{D}\mathbf{H})(s_{N_r})
\end{bmatrix}
\in\mathbb{R}^{(qN_r)\times h},
\qquad
\mathbf{A}_b :=
\begin{bmatrix}
(\mathcal{B}H_u)(\bar s_1)^\top\\
\vdots\\
(\mathcal{B}H_u)(\bar s_{N_b})^\top
\end{bmatrix}
\in\mathbb{R}^{N_b\times h},
\label{eq:oneshot_matrices_general}
\end{equation}
where $q$ is the state dimension (which depends on the system we aim to solve; here,
$q=2$ for the first-order diffusion system \eqref{eq:first_order_system}).
The corresponding targets are stacked as
\[
\mathbf{f}^\ast := \big[\mathbf{f}^\ast(s_n)\big]_{n=1}^{N_r}\in\mathbb{R}^{qN_r},
\qquad
\mathbf{b}^\ast := \big[b^\ast(\bar s_n)\big]_{n=1}^{N_b}\in\mathbb{R}^{N_b}.
\]
These targets encode the forcing and constraint data for the specific linear instance being solved.
For system \eqref{eq:first_order_system},
$\mathbf{f}^\ast(s)=[g(s),0]^\top$
contains the scalar forcing $g$ in the primary equation and a zero target for the auxiliary constraint.

\subsubsection{Algorithmic summary}\label{app:alg_details}

\begin{algorithm}[H]
\caption{Chebyshev-augmented one-shot perturbative PINN (offline/online pipeline)}
\label{alg:cheb_oneshot_ppinn}
\footnotesize
\begin{algorithmic}[1]
\Require Dominant linear operator $\mathcal{D}$ and constraint operator $\mathcal{B}$; sampling sets $\{s_n\}_{n=1}^{N_r}\subset\Omega$, $\{\bar s_n\}_{n=1}^{N_b}\subset\partial\Omega$; weights $w_{\mathrm{pde}},w_{\mathrm{bc}}>0$; Chebyshev degree $m$, quadrature size $M$; perturbation order $p$.
\Require Nonlinear instance $(f(\cdot;\eta),b(\cdot;\eta))$ and target $\varepsilon$.

\Statex \textbf{Offline: multi-head pretraining and precomputation.}
\State Train the multi-head model \eqref{eq:mh_model_vec} by minimizing \eqref{eq:mh_training_objective_updated}; freeze $\mathbf{H}(s)=\mathbf{H}_\theta(s)$.
\State Assemble fixed matrices $\mathbf{A}_r,\mathbf{A}_b$ from \eqref{eq:oneshot_matrices_general}.
\State Form $\mathbf{M}$ using \eqref{eq:M_q_updated} and store $\mathbf{M}^{-1}$.

\Statex \textbf{Online: Chebyshev surrogate + perturbative one-shot solves.}
\State Choose $[u_{\min},u_{\max}]$ and compute Chebyshev coefficients $\{c_\ell\}_{\ell=0}^m$ via \eqref{eq:c0_gc}--\eqref{eq:cl_gc}.
\State Initialize order-$0$ targets from \eqref{eq:order0} and form $\mathbf{q}^{(0)}$ via \eqref{eq:M_q_updated}; set $W_0=\mathbf{M}^{-1}\mathbf{q}^{(0)}$.
\For{$j=1$ \textbf{to} $p$}
  \State Build the forcing term for order $j$ using the Chebyshev lifted recurrence \eqref{eq:tau_def}--\eqref{eq:Gj_def} and the order equations \eqref{eq:orderj}.
  \State Form $\mathbf{q}^{(j)}$ from the stacked targets $(\mathbf{f}^{(j)},\mathbf{b}^{(j)})$ via \eqref{eq:M_q_updated} and compute $W_j=\mathbf{M}^{-1}\mathbf{q}^{(j)}$.
\EndFor
\State Reconstruct $u(s;\varepsilon)=\sum_{j=0}^{p}\varepsilon^j H_u(s)^\top W_j$.
\end{algorithmic}
\normalsize
\end{algorithm}

\subsection{Experimental details}\label{app:exp_details}

This section collects implementation details, architectures, sampling, and metric definitions referenced in
Section~\ref{sec:experiments}.

\subsubsection{Metrics}\label{app:metrics_details}

\paragraph{ODE residual MSE.}
For each ODE instance, the equation residual is evaluated on $N_t=100$ uniformly spaced points in $[0,5]$.
We report the mean squared residual, and then average over $100$ test instances.

\paragraph{PDE solution MSE.}
For PDE1 we evaluate the pointwise squared error against the manufactured solution on a fixed $61\times 61$
$(x,t)$ grid and average over the grid; the reported value is averaged across the full test suite.

\subsubsection{Chebyshev surrogate settings}\label{app:chebyshev_details}

All surrogates use Gauss--Chebyshev quadrature with $M=1000$ nodes.

\paragraph{ODE1.}
We approximate $\cos(u)$ on $[u_{\min},u_{\max}]=[-4.0,4.0]$ with degree $m=20$.

\paragraph{ODE2.}
We approximate $u^{-2}$ on $[u_{\min},u_{\max}]=[0.5,6.0]$ with degree $m=20$.

\paragraph{PDE1.}
We approximate $\frac{u}{u+1}-\delta u$ on $[u_{\min},u_{\max}]=[-0.9,1.0]$ with degree $m=30$.

\subsubsection{Constraint splitting across perturbation orders}\label{app:constraint_split_details}

We use an even split of constant boundary/initial constraints across perturbation orders. If the target
constraint value is $b$, then each order solves with
\[
b_{\mathrm{each}}=\frac{b}{\sum_{j=0}^{p}\varepsilon^j},
\]
so that reconstructing $u^{(p)}(s;\varepsilon)=\sum_{j=0}^p \varepsilon^j u_j(s)$ satisfies the original
constraint at the target $\varepsilon$.

\subsubsection{ODE architectures and training settings}\label{app:ode_arch_details}

Both ODE benchmarks (ODE1--ODE2) use the same multi-head network body with Sigmoid Linear Unit (SiLU) activations and $K=10$ heads.
The optimizer is Adam with learning rate $4\times 10^{-4}$ and a StepLR scheduler (step size $100$, $\gamma=0.92$).
Loss weights are $\alpha_{\mathrm{ode}}=0.5$ and $\alpha_{\mathrm{ic}}=1.5$. ODE1 is trained for $5000$ iterations
and ODE2 for $8000$ iterations, each using $50$ training points on $[0,5]$.

\paragraph{Train/test distributions.}
The full parameter ranges, initial-condition ranges, and test-suite sizes are those listed in the main text
(Section~\ref{sec:experiments}) and in the project configuration.

\subsubsection{PDE1 architecture and training settings}\label{app:pde1_details}

\paragraph{First-order system.}
To enforce the diffusion operator, we introduce $y=u_x$ and enforce a first-order system at interior points, with
boundary/initial conditions applied to $u$ only.

\paragraph{Network and optimization.}
The PDE network takes $(x,t)\in\mathbb{R}^2$ as input, uses SiLU activations, and has $K=16$ heads. Training
uses Adam with initial learning rate $10^{-3}$ and StepLR decay (step size $100$, $\gamma=0.98$) for $20000$
iterations. Interior sampling uses a $50\times 50$ grid; boundary sampling uses $100$ points on $t=0$ and $100$
points on each spatial boundary $x\in\{0,1\}$. Loss weights are
$w_{\mathrm{pde}}=w_{\mathrm{bc}}=w_{\mathrm{data}}=1$.

\paragraph{Manufactured solutions.}
Offline heads use $u(x,t)=A\sin(2\pi x)\sin(k\pi t)+b$ with $A\in\{0.5,-0.5\}$ and $k$ taking $8$ evenly spaced
values in $[1,2]$, yielding $16$ tasks. The test set uses
$u(x,t)=A\,x(x-1)\sin(kt)+b$ with $A\in\{1,2\}$, $k\in\{\pi,2\pi,3\pi,4\pi\}$, and $b\in\{0.2,0.4,0.6,0.8\}$,
and defines $f(x,t)$ by substitution into \eqref{eq:exp_pde1}.

\subsubsection{Timing protocol}\label{app:timing_details}

Reported online times include surrogate construction and the sequential computation of orders
$\{u_j\}_{j=0}^{p}$, and exclude offline training and the one-time precomputation of $\mathbf{M}^{-1}$.

\subsubsection{Gradient descent baseline}\label{app:head_gd_details}

\paragraph{Frozen trunk and head parameterization.}
This baseline employs the same pretrained multi-head backbone as the proposed method and freezes all shared-body parameters.
At test time, we optimize only the final linear head parameters $W$ that map the frozen features to the solution state.
Thus, this baseline measures the cost of iterative per-instance adaptation while keeping the representation fixed, in contrast to our closed-form one-shot head solve.

\paragraph{Objective.}
For ODE1--ODE2, we minimizes a physics-informed objective consisting of the ODE residual loss (enforcing the first-order system) plus an initial-condition (IC) loss.
For PDE1, we minimizes the PDE residual loss on interior points plus a boundary/initial-condition loss on constraint points.
All baseline runs use the same nonlinear instances (test suite) as used for evaluating our one-shot method.

\paragraph{Optimization, scheduling, and stopping.}
We use Adam with a StepLR schedule.
For ODE2 and PDE1, we additionally apply gradient clipping with max-norm $1.0$ to improve stability.
We employ an early-stopping criterion based on the mean-squared residual of the primary differential equation, with a maximum iteration cap.
If the early-stopping criterion is not met within the cap, the run is marked as reaching the maximum iteration limit.

\paragraph{Timing protocol.}
Baseline online time includes the full per-instance head optimization loop until early stopping or the iteration cap.
Reported runtimes are averaged across test instances that terminate before the cap.
As with our method, all online times exclude offline multi-head training and any one-time precomputations.

\begin{table}[H]
\caption{Baseline: optimization and stopping (CPU).}
\label{tab:head_gd_optstop}
\centering
\small
\begin{tabular}{lccc}
\hline
\textbf{Benchmark} & \textbf{Adam LR} & \textbf{StepLR (step,$\gamma$)} & \textbf{Stop / cap} \\
\hline
ODE1 & $10^{-2}$ & $(100,\,0.92)$ & $5\times10^{-4}$ / $2\!\times\!10^4$ \\
ODE2 & $10^{-2}$ & $(100,\,0.92)$ & $5\times10^{-3}$ / $2\!\times\!10^4$ \\
PDE1 & $3\!\times\!10^{-3}$ & $(200,\,0.96)$ & $1\times10^{-2}$ / $4\!\times\!10^3$ \\
\hline
\end{tabular}
\end{table}

\begin{table}[H]
\caption{Baseline: sampling and loss weights.}
\label{tab:head_gd_sampling}
\centering
\small
\begin{tabular}{lccc}
\hline
\textbf{Benchmark} & \textbf{Points per iter} & \textbf{Weights} & \textbf{Stabilization / init} \\
\hline
ODE1 & $N_t=100$ & $(w_{\rm ode},w_{\rm ic})=(0.5,1.5)$ & none / zero \\
ODE2 & $N_t=100$ & $(w_{\rm ode},w_{\rm ic})=(0.5,1.5)$ & clip($1.0$) / random \\
PDE1 & $I=60$, $B=200$ & $(w_{\rm pde},w_{\rm bc})=(1,1)$ & clip($1.0$) / zero \\
\hline
\end{tabular}
\end{table}

\subsection{Ablation study}\label{app:ablations}
We report ablations over perturbation strength $\varepsilon$, perturbation order $p$, and Chebyshev degree $m$ for ODE1 and PDE1. Unless otherwise specified, all other online settings follow the default configurations used in Table~\ref{tab:exp_summary}, and MSE is computed using the same metrics described in Appendix~\ref{app:metrics_details}.

For ODE1, all ablation studies are evaluated on the first parameter/initial-condition instance drawn from the same random test-generation procedure used in the main experiments.
For PDE1, all ablation studies are evaluated on a fixed manufactured instance with parameters $A=2$, $k=2\pi$, and $b=0.4$ (as defined in Appendix~\ref{app:pde1_details}), with all other settings unchanged.

\begin{table}[H]
\caption{ODE1 ablation over perturbation strength $\varepsilon$.}
\label{tab:ablate_ode1_eps}
\centering
\small
\begin{tabular}{cc}
\hline
$\varepsilon$ & MSE \\
\hline
0.05 & $3.40\times 10^{-8}$ \\
0.10 & $1.53\times 10^{-6}$ \\
0.20 & $7.94\times 10^{-7}$ \\
0.50 & $2.56\times 10^{-6}$ \\
0.80 & $2.74\times 10^{-5}$ \\
\hline
\end{tabular}
\end{table}

\begin{table}[H]
\caption{ODE1 ablation over perturbation order $p$.}
\label{tab:ablate_ode1_p}
\centering
\small
\begin{tabular}{cc}
\hline
$p$ & MSE \\
\hline
1  & $6.64\times 10^{-4}$ \\
2  & $9.97\times 10^{-5}$ \\
3  & $4.06\times 10^{-5}$ \\
4  & $2.63\times 10^{-5}$ \\
5  & $5.44\times 10^{-6}$ \\
6  & $2.91\times 10^{-6}$ \\
7  & $7.55\times 10^{-6}$ \\
8  & $1.75\times 10^{-6}$ \\
9  & $2.51\times 10^{-6}$ \\
10 & $1.80\times 10^{-7}$ \\
11 & $5.73\times 10^{-7}$ \\
12 & $2.56\times 10^{-6}$ \\
13 & $2.46\times 10^{-6}$ \\
14 & $1.51\times 10^{-5}$ \\
15 & $8.75\times 10^{-6}$ \\
16 & $3.82\times 10^{-7}$ \\
17 & $2.58\times 10^{-6}$ \\
18 & $5.64\times 10^{-6}$ \\
19 & $3.82\times 10^{-7}$ \\
20 & $2.38\times 10^{-7}$ \\
\hline
\end{tabular}
\end{table}

\begin{table}[H]
\caption{ODE1 ablation over Chebyshev degree $m$.}
\label{tab:ablate_ode1_m}
\centering
\small
\begin{tabular}{cc}
\hline
$m$ & MSE \\
\hline
1  & $4.52\times 10^{-1}$ \\
2  & $1.01\times 10^{-1}$ \\
3  & $1.01\times 10^{-1}$ \\
4  & $2.54\times 10^{-3}$ \\
5  & $2.54\times 10^{-3}$ \\
6  & $1.23\times 10^{-5}$ \\
7  & $1.23\times 10^{-5}$ \\
8  & $4.92\times 10^{-7}$ \\
9  & $4.84\times 10^{-7}$ \\
10 & $1.11\times 10^{-7}$ \\
11 & $1.22\times 10^{-7}$ \\
12 & $2.46\times 10^{-5}$ \\
13 & $3.54\times 10^{-5}$ \\
14 & $7.71\times 10^{-6}$ \\
15 & $1.46\times 10^{-5}$ \\
16 & $9.23\times 10^{-7}$ \\
17 & $9.06\times 10^{-7}$ \\
18 & $6.67\times 10^{-6}$ \\
19 & $6.58\times 10^{-6}$ \\
20 & $2.56\times 10^{-6}$ \\
\hline
\end{tabular}
\end{table}

\begin{table}[H]
\caption{PDE1 ablation over perturbation strength $\varepsilon$.}
\label{tab:ablate_pde1_eps}
\centering
\small
\begin{tabular}{cc}
\hline
$\varepsilon$ & MSE \\
\hline
0.05 & $3.84\times 10^{-6}$ \\
0.10 & $3.94\times 10^{-6}$ \\
0.20 & $4.66\times 10^{-6}$ \\
0.50 & $2.91\times 10^{-5}$ \\
0.80 & $2.41\times 10^{-4}$ \\
\hline
\end{tabular}
\end{table}

\begin{table}[H]
\caption{PDE1 ablation over perturbation order $p$.}
\label{tab:ablate_pde1_p}
\centering
\small
\begin{tabular}{cc}
\hline
$p$ & MSE \\
\hline
1  & $1.42\times 10^{-5}$ \\
2  & $4.00\times 10^{-5}$ \\
3  & $3.63\times 10^{-5}$ \\
4  & $3.23\times 10^{-5}$ \\
5  & $3.00\times 10^{-5}$ \\
6  & $2.92\times 10^{-5}$ \\
7  & $2.89\times 10^{-5}$ \\
8  & $2.88\times 10^{-5}$ \\
9  & $2.89\times 10^{-5}$ \\
10 & $2.90\times 10^{-5}$ \\
11 & $2.91\times 10^{-5}$ \\
12 & $2.91\times 10^{-5}$ \\
13 & $2.91\times 10^{-5}$ \\
14 & $2.91\times 10^{-5}$ \\
15 & $2.91\times 10^{-5}$ \\
16 & $2.91\times 10^{-5}$ \\
17 & $2.91\times 10^{-5}$ \\
18 & $2.91\times 10^{-5}$ \\
19 & $2.91\times 10^{-5}$ \\
20 & $2.91\times 10^{-5}$ \\
\hline
\end{tabular}
\end{table}

\begin{table}[H]
\caption{PDE1 ablation over Chebyshev degree $m$.}
\label{tab:ablate_pde1_m}
\centering
\small
\begin{tabular}{cc}
\hline
$m$ & MSE \\
\hline
2  & $4.10\times 10^{-4}$ \\
4  & $1.86\times 10^{-4}$ \\
6  & $8.50\times 10^{-5}$ \\
8  & $2.75\times 10^{-5}$ \\
10 & $3.50\times 10^{-5}$ \\
12 & $2.75\times 10^{-5}$ \\
14 & $2.93\times 10^{-5}$ \\
16 & $2.92\times 10^{-5}$ \\
18 & $2.90\times 10^{-5}$ \\
20 & $2.91\times 10^{-5}$ \\
22 & $2.91\times 10^{-5}$ \\
24 & $2.91\times 10^{-5}$ \\
26 & $2.91\times 10^{-5}$ \\
28 & $2.91\times 10^{-5}$ \\
30 & $2.91\times 10^{-5}$ \\
32 & $2.91\times 10^{-5}$ \\
34 & $2.91\times 10^{-5}$ \\
36 & $2.91\times 10^{-5}$ \\
38 & $2.91\times 10^{-5}$ \\
40 & $2.91\times 10^{-5}$ \\
\hline
\end{tabular}
\end{table}

\end{document}